\documentclass[letterpaper, 10 pt, conference, pdftex]{ieeeconf}  % Comment this line out if you need a4paper

\IEEEoverridecommandlockouts                              % This command is only needed if 
\usepackage{graphics} % for pdf, bitmapped graphics files
\usepackage{tabularx}
\usepackage{algorithm}
\usepackage{algpseudocode}
\usepackage{amsmath} % assumes amsmath package installed
\usepackage{amssymb}  % assumes amsmath package installed
\usepackage{bm}
\usepackage{booktabs} % For better table lines
\renewcommand{\theenumi}{\roman{enumi}}
\usepackage{notoccite} %表の置き場所によって引用の番号が大きくなってしまう問題への対処用

\usepackage{fancyhdr} % For IEEE 著作権表示用

\title{\LARGE \bf
PLEXUS Hand: Lightweight Four-Motor Prosthetic Hand\\ Enabling Precision--Lateral Dexterous Manipulation
}

\author{Yuki Kuroda$^{1}$, Tomoya Takahashi$^{1}$, Cristian C. Beltran-Hernandez$^{1}$, \\ Masashi Hamaya$^{1}$ and Kazutoshi Tanaka$^{1}$% <-this % stops a space
\thanks{$^{1}$OMRON SINIC X Corporation, Tokyo, Japan. 
        {\tt\small yuki.kuroda@sinicx.com}}%
}

\usepackage[dvipdfmx]{graphicx}
 \usepackage{multirow}
 \usepackage[table,xcdraw]{xcolor}
 \usepackage{gensymb}
\usepackage{amsmath}
\usepackage{mathcomp}
\usepackage{mathtools}

\usepackage[hang,small,bf]{caption}
\usepackage[subrefformat=parens]{subcaption}
\newcommand{\argmax}{\mathop{\rm arg~max}\limits}

\usepackage{boldline}

\newcommand{\Hline}{\hlineB{4}}
\fancypagestyle{plain}{
  \fancyhf{}
  
  \fancyfoot[L]{\parbox[t]{\textwidth}{
    \footnotesize © 2025 IEEE. Personal use of this material is permitted.  Permission from IEEE must be obtained for all other uses, in any current or future media, including reprinting/republishing this material for advertising or promotional purposes, creating new collective works, for resale or redistribution to servers or lists, or reuse of any copyrighted component of this work in other works.
  }}
}
\begin{document}

\maketitle
\thispagestyle{plain} % maketitle によってリセットされるスタイルを再度適用
%\thispagestyle{empty}
%\pagestyle{empty}

%%%%%%%%%%%%%%%%%%%%%%%%%%%%%%%%%%%%%%%%%%%%%%%%%%%%%%%%%%%%%%%%%%%%%%%%%%%%%%%%
\begin{abstract}
Electric prosthetic hands should be lightweight to decrease the burden on the user, shaped like human hands for cosmetic purposes, and have motors inside to protect them from damage and dirt. In addition to the ability to perform daily activities, these features are essential for everyday use of the hand. In-hand manipulation is necessary to perform daily activities such as transitioning between different postures, particularly through rotational movements, such as reorienting cards before slot insertion and operating tools such as screwdrivers. However, currently used electric prosthetic hands only achieve static grasp postures, and existing manipulation approaches require either many motors, which makes the prosthesis heavy for daily use in the hand, or complex mechanisms that demand a large internal space and force external motor placement, complicating attachment and exposing the components to damage. Alternatively, we combine a single-axis thumb and optimized thumb positioning to achieve basic posture and in-hand manipulation, that is, the reorientation between precision and lateral grasps, using only four motors in a lightweight (311~g) prosthetic hand. Experimental validation using primitive objects of various widths (5--30~mm) and shapes (cylinders and prisms) resulted in success rates of 90--100\% for reorientation tasks. The hand performed seal stamping and USB device insertion, as well as rotation to operate a screwdriver.
\end{abstract}

%%%%%%%%%%%%%%%%%%%%%%%%%%%%%%%%%%%%%%%%%%%%%%%%%%%%%%%%%%%%%%%%%%%%%%%%%%%%%%%%
\section{INTRODUCTION}\label{sec:Introduction}

Electric prosthetic hands are used by people who cannot perform hand functions because of congenital conditions or acquired causes such as accidents. These prostheses should be shaped similar to a biological human hand for cosmetic purposes, be lightweight (maximum acceptable weight is around 500~g~\cite{Vinet1995}) to decrease the burden on the user, have motors inside the hand due to protect them from damage and dirt, and be capable of performing activities of daily living (ADLs)~\cite{Carroll2004, Mohammadi2020, BelterDollar2011}. ADLs include fundamental tasks necessary for daily self-care, including housework and communication, which involve grasping and manipulating various objects such as tableware, writing implements, cash, and tools. As currently used prosthetic hands can only provide grasping functions~\cite{Belter2013, OttobockUS2023, Ossur2023}, we focus on incorporating in-hand manipulation capabilities to move objects between different grasping postures.

As the complex movements for in-hand manipulation require multiple motors, maintaining a lightweight design for daily use is challenging. A modular prosthetic limb is a sophisticated prosthetic hand that can perform in-hand manipulation~\cite{Johannes2011}. Although its 10 motors afford this capability, its weight of 1315~g makes it too heavy for daily use.

\begin{figure}[t]
  \centering
  \includegraphics[width=\linewidth]{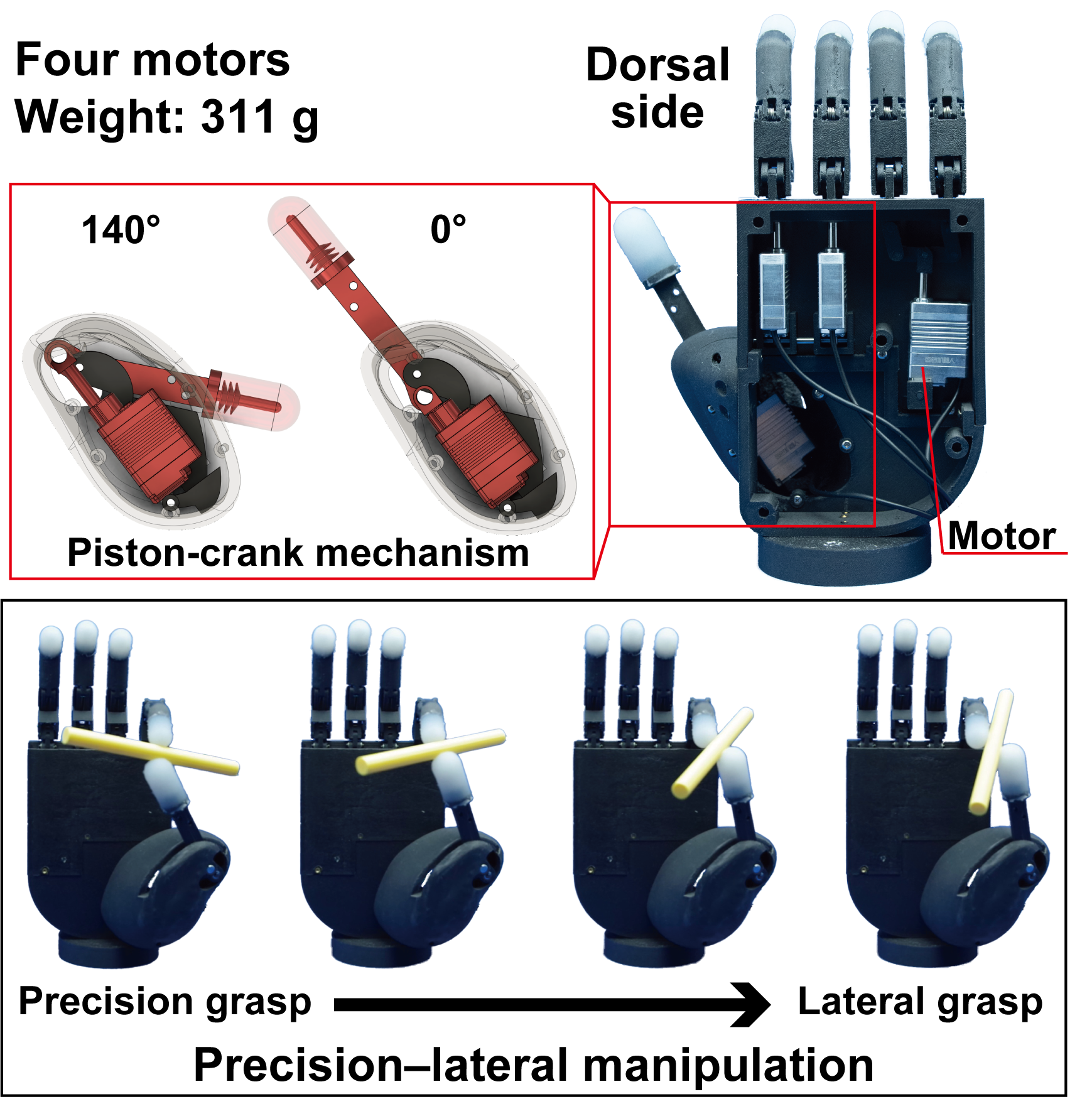}
  \caption{Configuration of PLEXUS hand and images of precision--lateral manipulation.}
  \label{fig:teaser}
\end{figure}

%三段落目 

%%%%%%%%%%%%%%%%%%%%%%%%%%%%%%%%%%%%

% ここから提案手法の話

In this study, we develop an electric prosthetic hand that can perform in-hand manipulation and basic hand postures with a minimal number of motors (NoM) for a lightweight design. Our approach enables rotational movements between different grasp types. To this end, we focus on the transition between precision and lateral grasps, which we refer to as precision--lateral manipulation, as illustrated in Fig.~\ref{fig:teaser}. This rotational capability enables performing essential daily tasks (e.g., reorienting objects such as cards before slot insertion and operating screwdrivers and other tools that require rotational manipulation) while minimizing compensatory movements.

\nocite{Liu2021}
\begin{table*}[t]
\centering
%\small
\caption{Related works}
\label{Table:RelatedWorks}
\begin{tabularx}{\linewidth}{>{\centering\arraybackslash}m{0.3cm} lccccX} 
\\ \Hline
 & \textbf{Hand} & \textbf{Weight (g)} & \textbf{Motor position} & \textbf{NoM (thumb)} & \textbf{Five basic postures} & \textbf{Manipulation} \\  \Hline

    \multirow{10}{*}{\rotatebox{90}{Prosthetic hands}}
     & \textbf{PLEXUS hand (proposed)} & \textbf{311} & Interior & \textbf{4 (1)} & 5 &  \textbf{precision--lateral} \\
     & BIT Hand~\cite{Wang2022, Kuroda2022} & 330 & Interior & 5 (1) & 4 &  \\ %\hline
     & i-limb quantum~\cite{Ossur2023} & 454--628 & Interior & 6 (2) & 5 &  \\ %\hline
     & Bebionic~\cite{OttobockUS2023} & 495--529~\cite{Belter2013} & Interior & 5 (1) & 5 &  \\ %\hline
     & TUAT/Karlsruhe humanoid hand~\cite{Fukaya2000} & 125 & Interior & 1 & 4 &  \\ %\hline
     & F hand~\cite{Fukaya2017} & & Interior & 1 & 4 &  \\ %\hline
     & SoftHand Pro~\cite{Godfrey2018} & 520 & Interior & 1 & 4 &  \\ %\hline
     & Belter and Dollar 2013~\cite{BelterDollar2013} & 350 & Interior & 1 & 4 &  \\ %\hline
     & Hannes hand~\cite{Laffranchi2020} & 480 & Interior & 1 & 4 &  \\ %\hline
     & Modular prosthetic limb~\cite{Johannes2011} & 1315 & Interior & 10 (4) & 5 & rolling \\ \hline
     \multirow{5}{0.0cm}[-2em]{\rotatebox{90}{Robotic Hands}} 
     & Shadow dexterous hand~\cite{Shadow2024} & 4000 & Exterior & 20 (5) & 5 & precision--lateral
+ reposition\\ %\hline
     & Twendy-one hand~\cite{Funabashi2018} & & Interior & 13 (4) & 5 & precision--lateral
+ reposition\\ %\hline
     & Allegro hand~\cite{Or2016} & 1500 & Interior  & 16 (4) & 5 & precision--lateral + reposition \\ %\hline
     & EthoHand~\cite{Konnaris2016} &  & Exterior & 7 (3) & 5 & precision--lateral \\ %\hline
     & Proposal by Kontoudis et al.~\cite{Kontoudis2019} & 650 & Exterior & 4 (2) & 4 & slight rolling + slight reposition \\ \Hline
\end{tabularx}
\end{table*}

We combined a single-axis thumb design at the carpometacarpal (CM) joint with optimized positioning. Liu et al.~\cite{Liu2021} reported that, although the human thumb is anatomically capable of movement along more than five axes, it primarily rotates around a single axis at the CM joint during daily grasping tasks, particularly during distal grasping, such as when using fingertips. Inspired by these findings, we propose a prosthetic hand design that enables in-hand manipulation in addition to basic hand postures by optimizing the position of the CM joint rotation axis in the thumb. This optimization maximizes the range of object widths for precision--lateral manipulation while ensuring the capability of basic hand postures.

We evaluated the performance of the prosthetic hand by testing the rotation of cylinders and square prisms with widths ranging from 5 to 30~mm and other objects in daily life. The results indicated high success rates for rotating cylinders, prisms, cards, personal seals, spoons, and screwdrivers. Practical applications included using personal seals, inserting USB devices, and turning screws using a screwdriver, suggesting the high potential effectiveness of the proposed prosthetic hand in ADLs.

The main contribution of this study is the development of a lightweight (311 g) prosthetic hand that enables both basic hand postures and precision--lateral manipulation using only four motors.

\section{RELATED WORK}  \label{related_works}
Two categories of hands, that is, practical electric prosthetic hands and anthropomorphic robotic hands capable of in-hand manipulation, are discussed. Table~\ref{Table:RelatedWorks} summarizes related work. We analyze the characteristics of available solutions based on existing studies and demonstrations focusing on total NoM, NoM in the thumb, weight, motor placement, capabilities in five basic hand postures (i.e., power, precision, tripod, and lateral grasps, as well as index pointing)~\cite{Cipriani2008, Cipriani2010} and manipulation. Table~\ref{Table:RelatedWorks} defines slight rolling and repositioning as minimal in-hand manipulations that do not enable transitions between static grasp postures.

\subsection{Anthropomorphic Robotic Hands with the Capability of Daily Life In-Hand Manipulation}
In-hand manipulation can be achieved through two approaches: high-NoM designs (e.g., 13 motors~\cite{Funabashi2018}, 16 motors~\cite{Or2016}, and 20 motors~\cite{Shadow2024}) and mechanical solutions with fewer motors. 
Notable examples of the latter include EthoHand (seven motors)~\cite{Konnaris2016}, which employs a tendon-driven ball joint for the thumb to reproduce human-like movements, and the design by Kontoudis et al., which achieves thumb abduction, adduction, and flexion using a tendon-driven differential mechanism with only two motors~\cite{Kontoudis2019}. Although these mechanical solutions achieve in-hand manipulation with fewer actuators, their mechanisms occupy space on the palm, impeding internal motor placement, which is crucial to ensure durability and space efficiency when attached to the residual limb of the user~\cite{Mohammadi2020}. Consequently, existing solutions either became too heavy (exceeding 1500~g) because of the large NoM or required palm space for their mechanisms. Our prosthetic hand addresses these limitations with a four-motor design weighing 311~g, which can perform five basic postures and in-hand manipulation while maintaining all mechanisms inside the hand.

\subsection{Lightweight Prosthetic Hand with Low NoM}
A modular prosthetic limb (10 motors) was developed as a prosthetic hand and can perform object rotation. However, its large NoM makes it too heavy (1315~g) for typical prosthesis users~\cite{Johannes2011}. To date, several practical electrical prosthetic hands have been proposed. Among them, the two widely adopted electric prosthetic hands in clinical use are bebionic (five motors)~\cite{OttobockUS2023} and i-Limb (six motors)~\cite{Ossur2023}. Most practical hands weigh approximately 500 g, which is the minimum weight of prosthetic hands mentioned in previous studies~\cite{Vinet1995}. However, further weight reduction is necessary to minimize the burden on the users.

Recently, electric prosthetic hands that incorporate adaptive grasp mechanisms (e.g., differential mechanisms) have been developed for achieving a lighter weight while enabling diverse grasping motions with a low NoM. The TUAT/Karlsruhe humanoid hand, which combines compound four-bar linkages with differential mechanisms, achieves four basic hand postures and all 14 types of Kamakura’s grasp taxonomy with only one motor~\cite {Fukaya2000}. Several other studies have achieved coordinated finger movements using differential mechanisms, enabling various hand postures with minimal actuation~\cite{Fukaya2017}–\cite{Laffranchi2020}. %~\cite{Fukaya2017, Godfrey2018, Xu2015, BelterDollar2013, Controzzi2017, Laffranchi2020}.

Although the abovementioned developments represent a significant progress in static grasping, technical challenges related to in-hand manipulation persist because the mechanically linked adaptive mechanism that connects all finger movements is incompatible with the independent finger control required for object manipulation~\cite{Ciocarlie2009}. Existing advanced anthropomorphic hands require from two to four motors~\cite{Konnaris2016, Kontoudis2019} for thumb rotation and two motors~\cite{Shadow2024, Funabashi2018, Or2016} for each non-thumb finger for translational operations, thereby making implementation challenging in prosthetic hands that must be lightweight and have internal actuators. In contrast, our hand achieves in-hand manipulation with a single motor for the thumb by combining a single-axis thumb design at the CM joint with optimized positioning.

\begin{table}[t]
\centering
\small
\caption{Parameters of PLEXUS Hand.}
\label{Table:Parameters}
\begin{tabularx}{\linewidth}{Xp{2cm}} \Hline
    \multicolumn{2}{c}{Grasp range requirements} \\ \hline
    Radius for precision grasp [$R_{\text{Gmin}}$, $R_{\text{Gmax}}$] & [0, 60] mm \\
    Radius for lateral grasp [$R_{\text{Gmin}}$, $R_{\text{Gmax}}$] & [0, 30] mm \\
    Radius for tripod grasp [$R_{\text{Tmin}}$, $R_{\text{Tmax}}$] & [10, 80] mm \\
    Width for in-hand manipulation [$W_{\text{min}}$, $W_{\text{max}}$] & [0, 30] mm \\ \hline
    \multicolumn{2}{c}{Common geometric parameters} \\ \hline
    $\delta_m$ (maximum fingertip deformation) & 4.87 mm\\ 
    $\theta_{\text{min}}$ (minimum grasp angle) & $110\tcdegree$ \\
    $\alpha_{\text{perm}}$ (permitted contact angle) & $30\tcdegree$ \\ \Hline
\end{tabularx}
\end{table}

\section{HAND DESIGN OPTIMIZATION}  \label{sec:hand_design}

\subsection{Design Requirements}\label{subsec:DesignRequirements}
\subsubsection{Grasping Requirements}
Cipriani et al. claimed that a hand should be able to perform five basic postures: power, precision, tripod, and lateral grasps, as well as index pointing~\cite{Cipriani2008, Cipriani2010}. The range of graspable object widths is based on the inter-fingertip distance, and only fingertip-based grasping postures (i.e., precision, tripod, and lateral grasps) are considered because palm-based grasping (i.e., power grasp) has uncertain contact positions, hindering range definition. Unlike human-enveloping grasps, prosthetic hands achieve a power grasp by adding palm contact to fingertip-based grasping. Therefore, satisfying the geometric requirements for precision grasping ensures power grasping. Based on daily object grasping analysis~\cite{Feix2014} and considering contact position uncertainty, we set the grasp range to approximately three times the average grasping width, as indicated in Table~\ref{Table:Parameters}.

\begin{algorithm}
\caption{Optimize thumb configuration for grasping and manipulation}\label{algorithm:Optimize}
\begin{algorithmic}[t]
\Require Four-finger configuration, search ranges, grasp constraints, manipulation parameters
\Ensure Optimal thumb configuration $\omega_{opt}$ ($x$, $y$, $z$, roll, pitch, yaw)
\Function{FindOptimalThumbConfiguration}{}
    \State $\Omega_{valid} \gets \emptyset$ \Comment{Valid configurations}
    \For{each thumb configuration $\omega$ in search space}
        \If{\Call{IsValidGrasp}{$\omega$}}
            \State $\Omega_{valid} \gets \Omega_{valid} \cup \{\omega\}$
        \EndIf
    \EndFor
    
    \State $\omega_{opt} \gets \text{null}$, $w_{max} \gets 0$ \Comment{Best configuration and its range}
    \For{each $\omega \in \Omega_{valid}$}
        \State $W(\omega) \gets \Call{CalculateManipulationRange}{\omega}$
        \If{$|W(\omega)| > w_{max}$}
            \State $\omega_{opt} \gets \omega$, $w_{max} \gets |W(\omega)|$
        \EndIf
    \EndFor
    
    \State \Return $\omega_{opt}$
\EndFunction
\end{algorithmic}
\end{algorithm}
 
\subsubsection{In-Hand Manipulation Requirements}
The hand must handle flat and curved surfaces to accommodate the manipulation of objects with different geometries. Based on daily object-grasping analysis studies~\cite{Feix2014}, the average object width for lateral and precision grasps is approximately 20~mm. We set the requirement to 30~mm for providing a margin above this average to perform in-hand manipulation.

\subsubsection{Implementation Requirements}
Actuators should be placed inside the hand for durability and space limitations~\cite{Mohammadi2020}. In addition, actuation must enable independent control of the thumb, index, and middle fingers for basic postures, with a synergistic ring--little finger actuation for achieving a power grasp using a minimal NoM. The selected prosthetic hand dimensions are based on the 50th percentile of the Japanese anthropometric data~\cite{Kouchi2012}. Moreover, the prosthesis employs actuators equivalent to those in clinically validated hands~\cite{Wang2022, Kuroda2022} to ensure sufficient grasping force, and they should achieve a significantly lower weight (under 350~g, excluding electronics) than common electric prostheses~\cite{OttobockUS2023, Ossur2023}.

\subsection{Optimization Method}
We aim to maximize the range of precision--lateral manipulation through optimal thumb positioning to achieve better manipulation capabilities for common objects while satisfying the requirements described in Sec.~\ref{subsec:DesignRequirements}. A manual search is impractical because even small position errors can lead to manipulation failures. This challenge explains why conventional prosthetic hands, in which thumb positioning is predominantly based on anatomical structures~\cite{Andres2022}, struggle with precision--lateral manipulation. Hence, we employ optimization to determine the position using the parameters defined in Table~\ref{Table:Parameters}.

\begin{figure}[t]
\center{
\includegraphics[keepaspectratio, width=\linewidth]{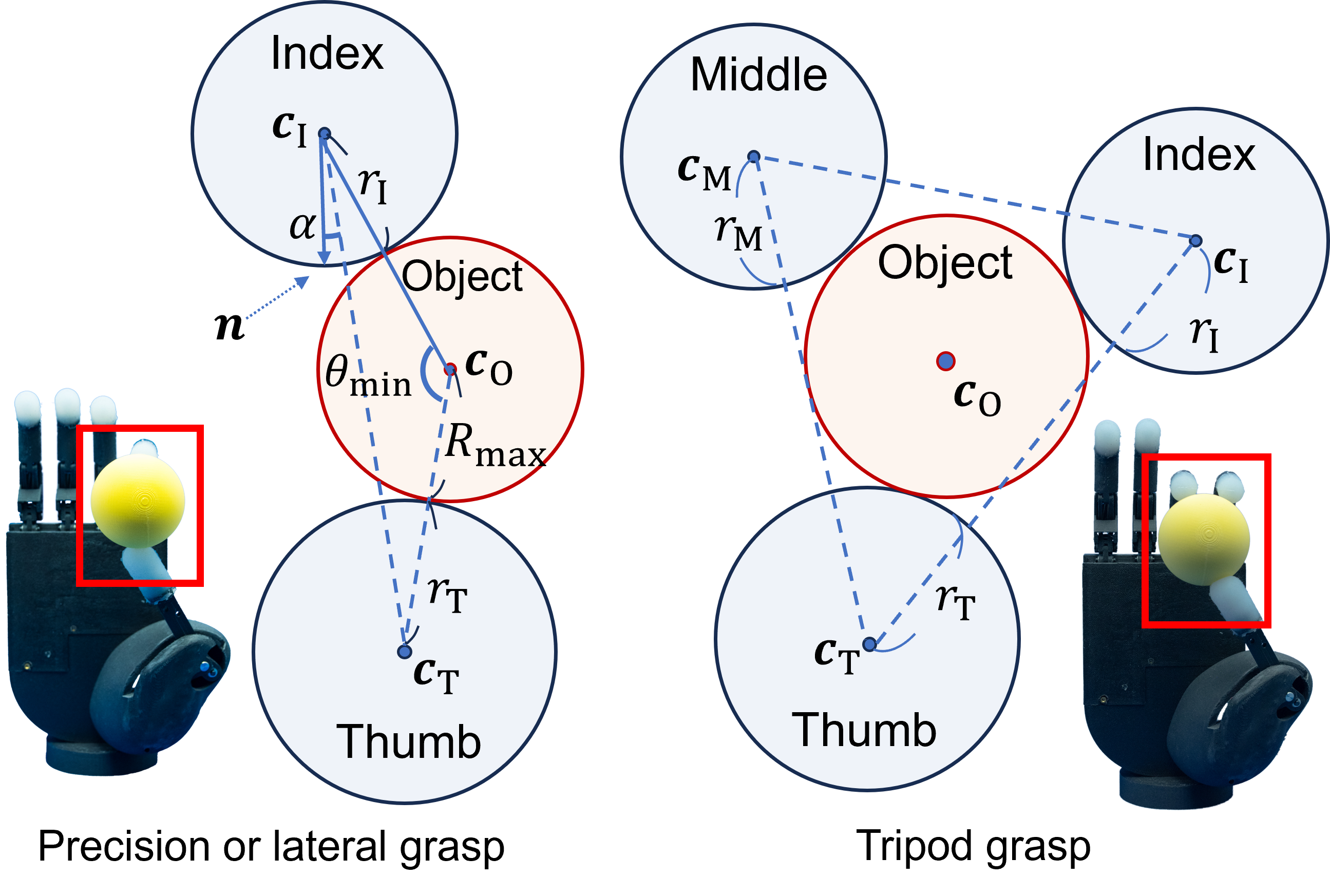} 
\caption{Geometric definitions of fingers and object to calculate grasp requirements. In precision and lateral grasps, the orientation of reference vector $\bm{n}$ differs. This vector defines the grasping direction of the object.}
\label{fig:GraspRequirements}
}
\end{figure}

We propose a two-stage optimization process to find the optimal thumb axis configuration, $\omega_{opt}$ ($x$, $y$, $z$, roll, pitch, and yaw). First, we identify configurations that satisfy all grasp requirements to reduce the search space and then select the configuration that enables the largest range of object widths for precision--lateral manipulation.

For simplicity, we model the fingertips and grasped objects as spheres. The grasped object is assumed to be sufficiently light such that its mass can be neglected. Let $\bm{c}_\mathrm{T}$, $\bm{c}_\mathrm{I}$, and $\bm{c}_\mathrm{M}$ represent the position vectors of the thumb, index, and middle fingers, respectively, and $\bm{c}_\mathrm{O}$ represent the position of the grasped object. Their radii are $r_\mathrm{T}$, $r_\mathrm{I}$, $r_\mathrm{M}$, and $R$.

The following sections describe the grasp requirements and manipulation range calculations implemented in \textsc{IsValidGrasp} and \textsc{CalculateManipulationRange}. The complete optimization process is summarized in Algorithm~\ref{algorithm:Optimize}.

\subsubsection{Grasp Validation}\label{subsubsec:GraspValidation}
Function \textsc{IsValidGrasp} in Algorithm~\ref{algorithm:Optimize} evaluates whether a given thumb configuration satisfies the requirements for precision, lateral, and tripod grasps. For thumb configuration $\omega$ to be valid, all three grasp types must be achievable within their respective object size ranges.

For precision and lateral grasps, we consider the relative positions between thumb position $\bm{c}_\mathrm{T}(\omega)$ determined by configuration parameters $\omega$ and index finger position $\bm{c}_\mathrm{I}$. A configuration is considered valid if it satisfies the following geometric range requirements:
\begin{itemize}
    \item The maximum graspable distance, $R_{\text{max}}$, calculated as shown in Fig.~\ref{fig:GraspRequirements}, must satisfy $R_{\text{max}} \geq R_{\text{Gmax}}$. This limit is determined by $\theta_{\text{min}}$ (illustrated in Fig.~\ref{fig:GraspRequirements}), which represents the minimum angle between vectors $\bm{c}_\mathrm{T}(\omega)-\bm{c}_\mathrm{O}$ and $\bm{c}_\mathrm{I}-\bm{c}_\mathrm{O}$ required for stable grasping considering friction.
    \item The minimum distance between fingertips, $R_{\text{min}}$, occurs when $\bm{c}_\mathrm{T}(\omega)$, $\bm{c}_\mathrm{I}$, and $\bm{c}_\mathrm{O}$ are colinear, and it must satisfy $R_{\text{min}} \leq R_{\text{Gmin}}$.
\end{itemize}

%TODO:Fig
\begin{figure*}[t]
  \centering
  \includegraphics[width=16cm]{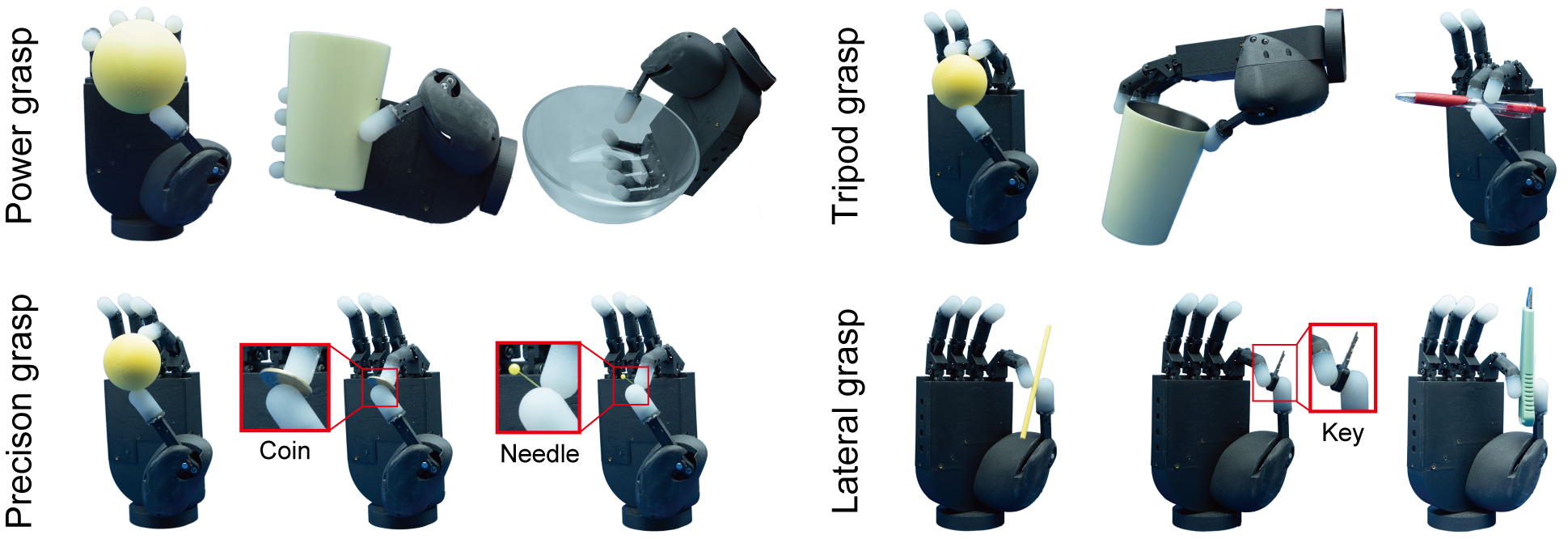}
  \caption{Execution of basic postures using PLEXUS hand. Index pointing was omitted because the index finger can clearly move independently.}
  \label{fig:BaseGrasps}
\end{figure*}

Grasping is established when the following conditions are satisfied:

\begin{itemize}
    \item As illustrated in Fig.~\ref{fig:GraspRequirements}, angle $\alpha$ between reference vector $\bm{n}$ (normal to the optimal contact surface of the index finger) and the grasp direction must satisfy $-\alpha_{\text{perm}} \leq \alpha \leq \alpha_{\text{perm}}$, where $\alpha_{\text{perm}}$ is empirically determined due to the lack of standardized angular definitions for grasp types.
    \item The force-applying digit (i.e., index finger for precision grasp and thumb for lateral grasp) must maintain its motion direction at an angle less than 45° relative to vector $\bm{c}_\mathrm{T}(\omega)-\bm{c}_\mathrm{I}$ and $\bm{c}_\mathrm{I}-\bm{c}_\mathrm{T}(\omega)$, respectively, to maintain the appropriate force direction.
\end{itemize}

For the tripod grasp, we evaluate the triangle formed by the thumb ($\bm{c}_\mathrm{T}(\omega)$), index finger ($\bm{c}_\mathrm{I}$), and middle finger ($\bm{c}_\mathrm{M}$). The object center, $\bm{c}_\mathrm{O}$, must lie within this triangle, and the achievable grasp radii, $[R_\text{min}, R_\text{max}]$, must encompass the specified range ($R_\text{min} \leq R_\text{Tmin}$ and $R_\text{max} \geq R_\text{Tmax}$).

%Note that while the physical minimum graspable thickness is 0 mm, we set $R_{\text{Gmin}}$ to -1 mm during optimization for numerical stability. This ensures reliable detection of valid configurations for grasping extremely thin objects.

\subsubsection{Manipulation Range Analysis}\label{subsubsec:ManipulationRangeAnalysis}
Function \textsc{CalculateManipulationRange} analyzes the precision--lateral transition path where the thumb moves from the lateral to the precision grasp position to determine the range of object sizes that can be manipulated. Let $i \in I$ and $j \in J$ be discretized steps along the motion ranges of the index finger and thumb, respectively. For each index finger position $\bm{c}_\mathrm{I}(i)$ and corresponding thumb position $\bm{c}_\mathrm{T}(\omega, j)$ determined by configuration $\omega$, we identify two critical points: $j_{i,\text{L}}\in J$ (i.e., the last position allowing lateral grasp, Sec.~\ref{subsubsec:GraspValidation}) and $j_{i,\text{P}}\in J$ (i.e., the first position permitting precision grasp, Sec.~\ref{subsubsec:GraspValidation}). The manipulation interval is $J_m = \{j \in J \mid j_{i,\text{L}} \leq j \leq j_{i,\text{P}}\}$.

For each index position $i$, the maximum and minimum distances during the transition are calculated as $d_{\text{max}}(\omega, i) = \max_{j \in J_m} |\bm{c}_\mathrm{I}(i) - \bm{c}_\mathrm{T}(\omega, j)|$ and $d_{\text{min}}(\omega, i) = \min_{j \in J_m} |\bm{c}_\mathrm{I}(i) - \bm{c}_\mathrm{T}(\omega, j)|$.

The range of manipulatable object widths $W_i(\omega)$ at each index position $i$ for thumb configuration $\omega$ is defined as
\begin{equation}
W_i(\omega) = [d_{\text{max}}(\omega, i) - (r_\mathrm{T} + r_\mathrm{I}), d_{\text{min}}(\omega, i) - (r_\mathrm{T} + r_\mathrm{I})+ 2\delta_m],
\end{equation}
where $\delta_m$ represents the allowable deformation of the flexible fingertip, calculated as $\delta_m = \sqrt{F/(\pi E)}$~\cite{Fakhari2015} with an assumed pinch force $F = 10~\mathrm{N}$ based on a prosthetic hand using equivalent actuators~\cite{Wang2022}, and Young's modulus $E$ given in Table~\ref{Table:Specifications}. The lower bound is determined by subtracting the fingertip radii from the maximum distance, whereas the upper bound considers the additional grasp range enabled by fingertip deformation.

The overall manipulation range, $W(\omega)$, for the configuration is calculated as the intersection of the ranges across the set of all index finger positions $I$.
\begin{equation}
W(\omega) = \bigcap_{i \in I} W_i(\omega)
\end{equation}
This ensures that objects within range $W(\omega)$ can be stably manipulated throughout the precision--lateral transition.

Among all valid thumb configurations $\Omega_{\text{valid}}$ found by \textsc{IsValidGrasp}, we select optimal configuration $\omega_{\text{opt}}$ that maximizes the manipulation range.
\begin{equation}
\omega_{\text{opt}} = \argmax_{\omega \in \Omega_{\text{valid}}} |W(\omega)|
\end{equation}
{\parskip=2mm
\section{PLEXUS Hand}\label{Section:prptotype}}
Based on the optimization described in Sec.~\ref{sec:hand_design}, we engineered a prototype of the PLEXUS hand (Precision--Lateral dEXteroUS manipulation hand), incorporating the optimal thumb axis position as determined by our optimization algorithms. The PLEXUS hand achieves both precision--lateral manipulation and the basic hand postures required for prosthetic applications (Fig.~\ref{fig:BaseGrasps}) while maintaining a simple design with only four motors, including a single motor for the thumb. Table~\ref{Table:Specifications} lists the specifications of the PLEXUS hand and optimization conditions.

\begin{figure}[t]
\center{
\includegraphics[keepaspectratio, width=\linewidth]{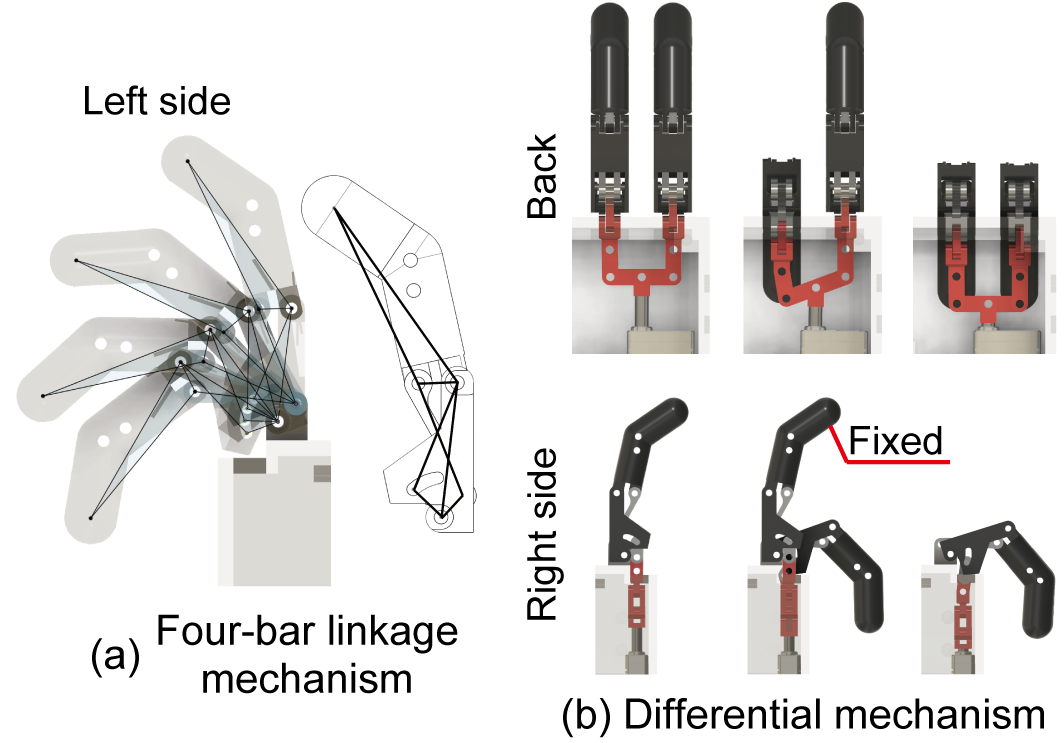} 
\caption{(a) Actuation and structure of four-bar linkage mechanism. (b) Differential mechanism that distributes motion between the fingers. When one finger is constrained, the displacement is redirected entirely to the unconstrained finger. Without constraints, motion is distributed to both fingers equally.}
\label{fig:Linkage}
}
\end{figure}

\begin{table}[t]
\centering
\small
\caption{Specifications of PLEXUS hand.}
\label{Table:Specifications}
\begin{tabularx}{\linewidth}{XX} \Hline
    \multicolumn{2}{c}{Physical specifications} \\ \hline
    Dimensions (W × D × H) & 130 × 30 × 210 mm \\
    Weight & 311 g \\
    NoM (thumb) & 4 (1) \\
    Main material & Nylon (3D printed) \\ \hline
    \multicolumn{2}{c}{Fingertip silicone} \\ \hline
    Type & TSG-E10 (TANAC, Japan)\\
    Young's modulus & 134.3 kPa~\cite{Yabuki2020}\\ \hline
    \multicolumn{2}{c}{Optimization} \\ \hline
    Optimization method & Grid search \\
    Number of axis positions explored & 19,200,000 \\ \Hline
 \end{tabularx}
 \end{table}

The PLEXUS hand features distinct actuation mechanisms for different digit groups. All digits are driven by linear actuators to simplify the control system integration.
\begin{enumerate}
    \item Thumb: Driven by a LAS16-023D linear motor (Beijing Inspire Robots Technology, China) and utilizing a piston--crank mechanism~\cite{Wang2022} as the simplest solution for linear-to-rotary motion conversion, as shown in Fig.~\ref{fig:teaser}.
    \item Index and middle fingers: Actuated via four-bar linkage mechanisms~\cite{Wang2022}, as shown in Fig.~\ref{fig:Linkage}(a). Each mechanism transforms the linear motion of LAS10-023D actuators (Beijing Inspire Robots Technology) into natural curling movements, enabling multiple contact points with objects for stable grasping.
    \item Ring and little fingers: Sharing a single LAS10-023D actuator through a differential mechanism~\cite{Fukaya2017}, as shown in Fig.~\ref{fig:Linkage}(b). The differential mechanism automatically distributes the displacement of the actuator between both fingers. When one finger encounters an object, the other can continue to move, enabling adaptive grasping while reducing the number of actuators needed.
\end{enumerate}

\section{EXPERIMENT AND RESULTS}\label{sec_experiment}
We conducted a quantitative evaluation using primitive and common objects to verify the capability of the proposed prosthetic hand to perform precision--lateral manipulation across a range of object sizes and shapes and a preliminary demonstration of practical applications for validating the effectiveness of the hand in real-world scenarios.

\begin{figure*}[t]
\center{
\includegraphics[width=\linewidth]{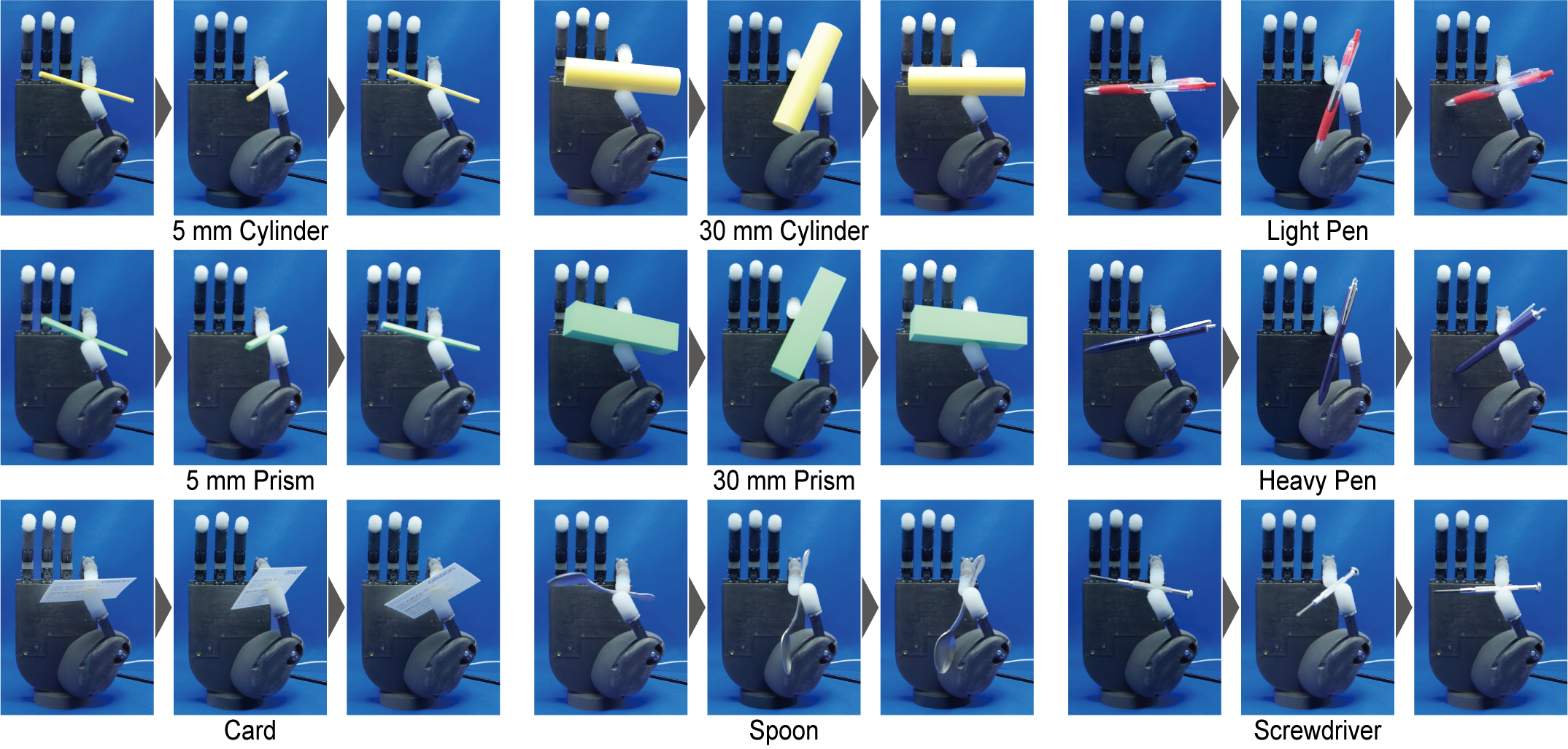} 
\caption{Precision--lateral manipulation experiments for each object.}
\label{fig:Experiment}
}
\end{figure*}

\subsection{Experiment 1: Quantitative Evaluation with Primitive and Common Objects}
\subsubsection{Objects Used}
The following two object categories were used for Experiment 1:
\begin{itemize}
	\item {\bf Primitive objects:} Square prisms and cylinders with height of 120 mm. Both types varied in width/diameter from 5--30 mm in 5 mm increments, totaling six variations per object. These objects allowed us to evaluate performance on flat and curved surfaces.
  	\item {\bf Common objects of daily living:} Business card, light pen, heavy pen, seal stamp, spoon, and precision screwdriver~\cite{Owsley2001}. These objects were chosen as representative examples of objects requiring precision--lateral manipulation in everyday situations.
\end{itemize}

\subsubsection{Experimental Procedure}
Sec.~\ref{subsubsec:ManipulationRangeAnalysis} determines the optimization target range between the thumb positions corresponding to trajectory elements $j_{i,\text{P}}$ and $j_{i,\text{L}}$ for object manipulation. We implemented the following setup to experimentally validate the manipulation capability in this range. Considering mechanical errors and calibration constraints, we prioritized achieving the target position for lateral grasp (corresponding to $j_{i,\text{L}}$) because precise positioning is crucial owing to its dependence on geometric constraints. For precision grasping, we adopted an initial position with the thumb directly below the index finger. The index finger position was adjusted for each object to enable secure holding in a precision grasp and remained fixed throughout manipulation, while only the thumb moved. This setup resulted in a validation over an extended range, including the position corresponding to $j_{i,\text{P}}$.

For each object, we performed the following steps 10 times:
\begin{enumerate}
\item The object was grasped near its center using a precision grasp, and its stability was manually confirmed by slightly moving the object.
\item The manipulation from precision to lateral grasp was executed over 2.5 s, with the thumb moving to the position corresponding to $j_{i,\text{L}}$ optimized for each object width.
\item The grasp was maintained at the lateral position for 2 s to confirm stability.
\item The manipulation from lateral back to precision grasp was executed over 2.5 s.
\end{enumerate}
Manipulation was considered successful if the object was grasped during both transitions.

\subsubsection{Results}
The success rates for precision--lateral manipulation are listed in Table~\ref{Table:SuccessRates}. Images of the experiment are shown in Fig.~\ref{fig:Experiment}.

 \begin{table}[t]
 \centering
 \small
 \caption{Success rates for precision--lateral manipulation.}
 \label{Table:SuccessRates}
 \begin{tabularx}{\linewidth}{cccc} 
 \Hline
  Width (mm)          & Primitive type & Weight (g) & Success rate (\%) \\ \Hline
\multirow{2}{*}{5}  & Cylinder       & 2.79 & 100                       \\
                    & Square prism   & 2.49 & 100                       \\ \hline
\multirow{2}{*}{10} & Cylinder       & 5.04 & 90                        \\
                    & Square prism   & 6.20 & 100                       \\ \hline
\multirow{2}{*}{15} & Cylinder       & 9.90 & 100                       \\
                    & Square prism   & 11.59 & 100                       \\ \hline
\multirow{2}{*}{20} & Cylinder       & 14.58 & 100                       \\
                    & Square prism   & 18.03 & 100                       \\ \hline
\multirow{2}{*}{25} & Cylinder       & 20.80 & 100                       \\
                    & Square prism   & 24.65 & 100                       \\ \hline
\multirow{2}{*}{30} & Cylinder       & 28.34 & 100                       \\
                    & Square prism   & 34.34 & 100                       \\ \hline
        0.3        & Card           & 1.16 & 100                       \\
        10.3       & Heavy pen      & 23.45 & 60                        \\
        10.2       & Light pen      & 9.75 & 50                        \\
        9.9        & Seal           & 4.68 & 100                       \\
        1.7        & Spoon          & 15.48 & 100                       \\
        6.2        & Screwdriver    & 21.22 & 80  \\ \Hline
\end{tabularx}
\end{table}

\begin{figure*}[t]
\center{%\textwidth
\includegraphics[width=16.0cm]{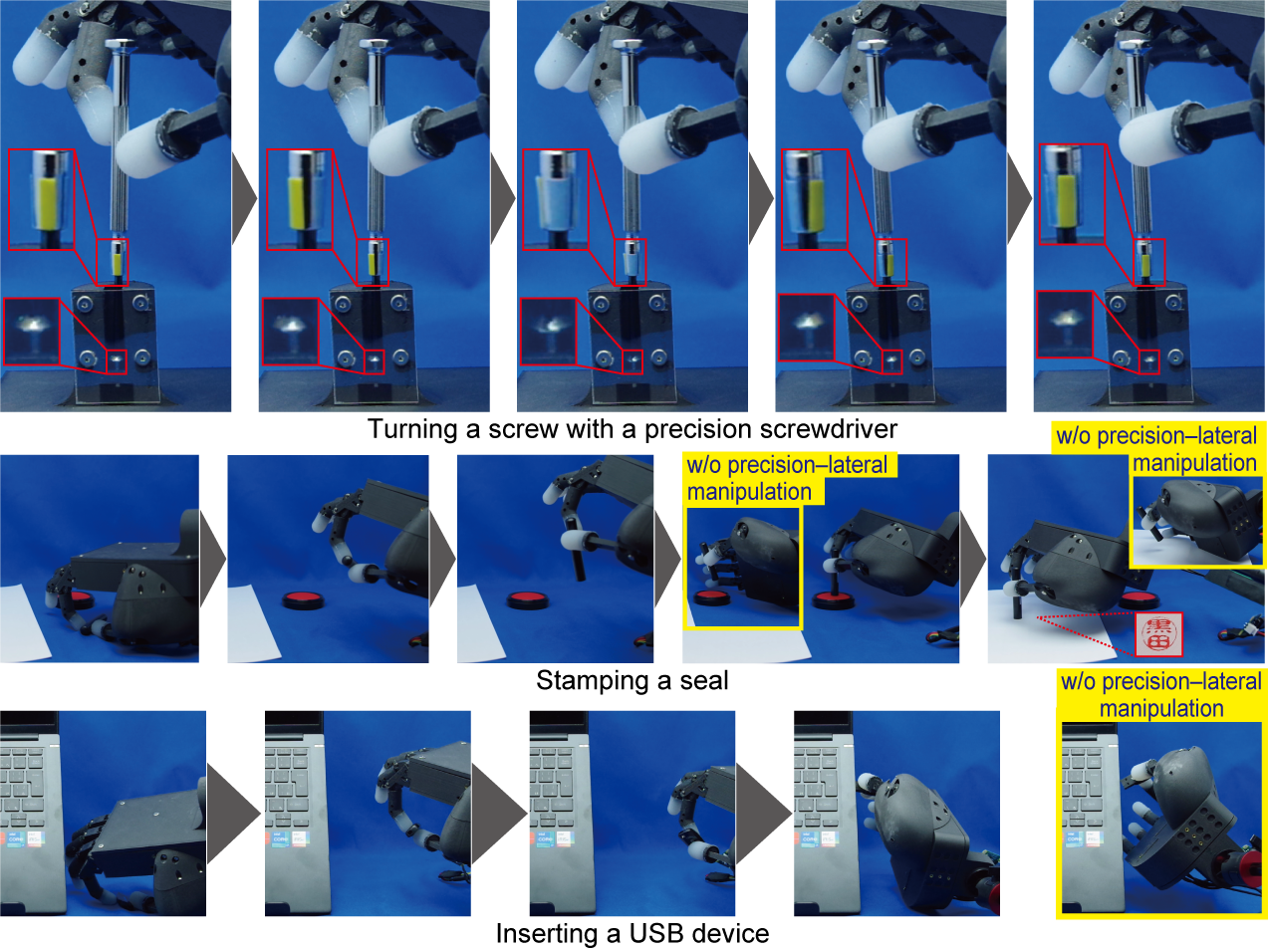} 
\caption{Precision--lateral manipulation and its applications. Without precision--lateral manipulation (yellow frame), the palm blocks the movement.}
\label{fig:PracticalUse}
}
\end{figure*}

\subsection{Experiment 2: Demonstration of Practical Applications}

\subsubsection{Experimental Setup}
Although comprehensive testing with users wearing a prosthetic hand remains a key objective for future work, we conducted preliminary demonstrations in three practical scenarios to illustrate the potential real-world applications of our approach and developed prosthesis.

\begin{enumerate}
    \item {\bf Precision screwdriver operation:} This task demonstrated turning screws with a precision screwdriver. The PLEXUS hand lacked the flexibility afforded by a human hand with a flexible palm for stabilization. This hindered steady handling of the screwdriver. Therefore, guides were used to maintain the position of the screwdriver. Developing a flexible palm for better object manipulation will be part of future work.
    \item {\bf Seal (hanko) usage:} We also demonstrated the ability to pick up, manipulate, and apply a seal to a document. This task addressed a common need in Japan, particularly for package receipt confirmation.
    \item {\bf USB device insertion:} Finally, we demonstrated the capability of the prosthetic hand to pick up, manipulate, and insert a USB device into a laptop port.
\end{enumerate}

\subsubsection{Results}
Images of the practical demonstrations are shown in Fig.~\ref{fig:PracticalUse}. The ``w/o precision--lateral manipulation'' condition refers to cases without precision--lateral manipulation, where palmar interference prevented successful task execution.

\section{DISCUSSION}
\subsection{Experimental Results}
\subsubsection{Primitive Objects}
Experiments with primitive objects generally achieved high success rates. However, narrower objects with smaller contact areas tended to be less stable, particularly cylinders. This instability resulted from the reduced elastic force and friction caused by a low compression of the flexible tissue of the fingertip on small cross-sectional areas.

\subsubsection{Common Objects of Daily Living}
Experiments with common objects of daily living revealed instability when handling heavy items relative to their width and cylindrical shape, such as a pen or screwdriver, making it difficult to secure an adequate contact area. The primary challenge was the insufficient pinch force to manipulate or hold heavy objects. This limitation stemmed from our fixed-finger control approach. Although precision grasps achieved a proper force direction where the thumb trajectory intersected the motion path of the index finger, lateral grasps were limited because the fixed index finger could not be positioned along the motion path of the thumb. Future improvements should involve repositioning the index finger during manipulation, which may require advanced control systems with touch sensors for adaptive adjustments. Furthermore, a spherical finger shape decreased the contact area with cylindrical objects, making finger shape optimization another challenge for future work.

\subsection{Practical Use}
Conventional practical prosthetic hands cannot be used to perform rotational movements essential for tasks such as turning screws. Furthermore, they often require non-disabled hands to reposition objects or force users to adopt physically demanding postures~\cite{Chadwell2018, Williams2021, Spiers2017}. Our developed prosthetic hand may increase the number of achievable ADLs while reducing the burden on the user by enabling in-hand manipulation using only the prosthetic hand.

\section{CONCLUSIONS}
We present a novel approach to design a four-motor lightweight electric prosthetic hand that can perform five basic hand postures and precision--lateral manipulation. Our PLEXUS hand demonstrates the ability to manipulate objects of various widths and common objects of daily living with high success rates. Our study and findings provide insights into improving the performance of electric prosthetic hands, potentially enhancing the quality of life of people with upper limb deficiencies.

%%%%%%%%%%%%%%%%%%%%%%%%%%%%%%%%%%%%%%%%%%%%%%%%%%%%%%%%%%%%%%%%%%%%%%%%%%%%%%%%

\section*{ACKNOWLEDGMENT}
This work was supported by JSPS KAKENHI Grant Number JP24KJ0248.

%%%%%%%%%%%%%%%%%%%%%%%%%%%%%%%%%%%%%%%%%%%%%%%%%%%%%%%%%%%%%%%%%%%%%%%%%%%%%%%%

\bibliographystyle{IEEEtran}
\bibliography{reference_20241126}

\end{document}